\documentclass[letterpaper, 10 pt, conference]{ieeeconf}  
\usepackage{graphicx} 
\usepackage{xcolor}
\usepackage{colortbl}
\usepackage{amsmath}
\definecolor{tabfirst}{RGB}{255, 153, 153}
\definecolor{tabsecond}{RGB}{255, 204, 153}
\definecolor{tabthird}{RGB}{255, 248, 174}
\IEEEoverridecommandlockouts   

\title{\LARGE \bf
Multi-view Normal and Distance Guidance Gaussian Splatting for Surface Reconstruction}

\author{Bo Jia$^{1}$, Yanan Guo$^{1}$, Ying Chang$^{2}$*, Benkui Zhang$^{2}$, Ying Xie$^{1}$, Kangning Du$^{1}$, Lin Cao$^{1}$
\thanks{*This work was supported in part by the National Natural Science Foundation of China under Grant U20A20163, Grant 62201066 and Grant 62001033, and in part by Beijing Municipal Education Commission Research Program under Grant KZ202111232049 and Grant KM202111232014. (Corresponding author: Ying Chang.)}
\thanks{$^{1}$The Key Laboratory of Information and Communication Systems, Ministry of Information Industry, Beijing Information Science and Technology University, Beijing 100192, China}
\thanks{$^{2}$The Key Laboratory of Target Cognition and Application Technology, Aerospace Information Research Institute, CAS, Beijing 100094, China}
}

\begin{document}

\maketitle
\thispagestyle{empty}
\pagestyle{empty}

\begin{abstract}

3D Gaussian Splatting (3DGS) achieves remarkable results in the field of surface reconstruction. However, when Gaussian normal vectors are aligned within the single-view projection plane, while the geometry appears reasonable in the current view, biases may emerge upon switching to nearby views. To address the distance and global matching challenges in multi-view scenes, we design multi-view normal and distance-guided Gaussian splatting. This method achieves geometric depth unification and high-accuracy reconstruction by constraining nearby depth maps and aligning 3D normals. 
Specifically, for the reconstruction of small indoor and outdoor scenes, we propose a multi-view distance reprojection regularization module that achieves multi-view Gaussian alignment by computing the distance loss between two nearby views and the same Gaussian surface. Additionally, we develop a multi-view normal enhancement module, which ensures consistency across views by matching the normals of pixel points in nearby views and calculating the loss. Extensive experimental results demonstrate that our method outperforms the baseline in both quantitative and qualitative evaluations, significantly enhancing the surface reconstruction capability of 3DGS. 
Our code will be made publicly available at (https://github.com/Bistu3DV/MND-GS/).


\end{abstract}

\section{INTRODUCTION}

Surface reconstruction is a challenging task in the fields of computer graphics and computer vision, with promising applications spanning heritage conservation \cite{her1, her2}, game development \cite{game1, game2}, architectural design \cite{arc2, arc3}, and clinical medicine \cite{med1, med2, med3}. Neural Radiance Fields (NeRF) \cite{NeRF} and its variants have made significant breakthroughs in surface reconstruction, praised for robustness and fidelity. However, NeRF suffers from high memory usage and long rendering times. To address this, 3D Gaussian Splatting (3DGS) \cite{3DGS} has been refined for real-time rendering. 3DGS uses Structure from Motion (SfM) to obtain depth and camera parameters from multi-view images, converting them to 3D point clouds. Based on these, 3D Gaussian distributions are initialized. During optimization, depth information enforces geometric consistency, adjusting distributions to fit the scene. Finally, depth is used for visibility and occlusion processing, enabling high-quality 3D reconstruction.

\begin{figure}[h]
    \centering
\includegraphics[width=1\linewidth]{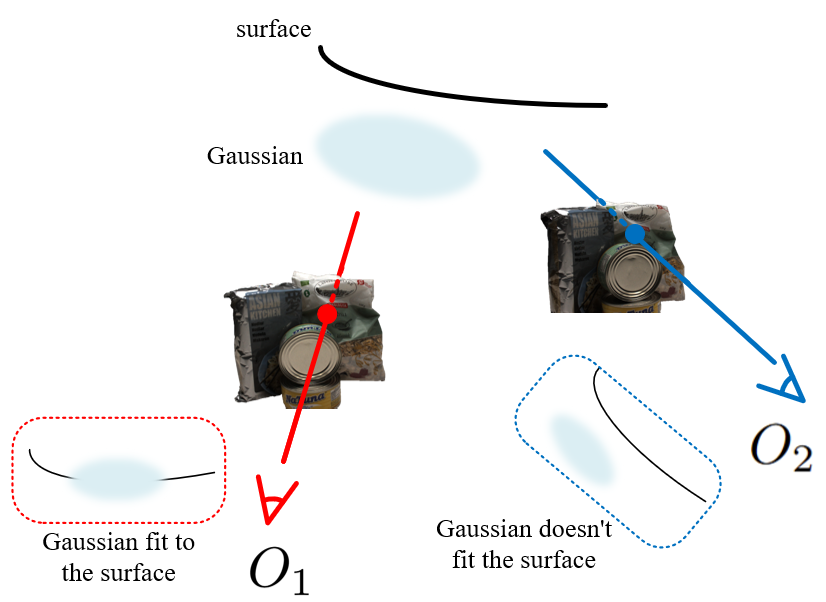}
\caption{Limitation in single-view geometric constraints. When viewed from the direction of viewpoint $O_1$, the Gaussian curve is adhered to the surface. However, when viewed from the direction of viewpoint $O_2$, the Gaussian curve is detached from the surface.}
\label{fig:intro}
\end{figure}

Currently, some methods \cite{PGSR,2DGS,RaDe-GS} use 3DGS with geometric constraints to integrate Gaussian distributions with scene surface features. These maintain real-time rendering while enhancing reconstruction accuracy and consistency through techniques like normal alignment. They focus on single-view constraints, while PGSR's \cite{PGSR} multi-view constraints address inconsistencies across views. However, relying solely on single-view constraints has limitations, as shown in Fig. \ref{fig:intro}. When Gaussian normals align with a single-view plane, their spatial distribution may misalign from nearby views. This is due to single-view optimization seeking local solutions, adjusting Gaussians excessively to align with the current view, losing global consistency.


To address these issues, we propose multi-view normal and distance guidance Gaussian splatting, which lies in applying global constraints from a multi-view perspective, ensuring consistency in scene geometry across different viewpoints.  Specifically, we design a multi-view distance reprojection regularization module to tackle the issue of inaccurate distance estimation in multi-view conditions. This module imposes constraints on the distance maps of two nearby views, effectively achieving geometric distance uniformity and enhancing the accuracy and realism of scene reconstruction. Additionally, the Gaussian face challenges in accurately fitting surfaces in a multi-view state, for which we develop a multi-view normal enhancement module. This module achieves view consistency by sampling the surrounding pixels of nearby pixel points, mapping them into 3D space, and computing a loss between the normals of the simulated plane. 

In this paper, our main contributions are summarized as follows: 1) We propose a multi-view distance reprojection regularization module. The core function of this module is to constrain nearby distance maps to ensure consistency in geometric distance, thereby addressing the issue of inaccurate distance estimation in multi-view conditions. 2) We propose an innovative multi-view normal enhancement module. This module aims to achieve precise alignment of normals between nearby views, significantly enhancing the robustness of normal estimation and ensuring high consistency of geometric information across different viewpoints. 3) We conduct experiments on DTU and Mip-NeRF360 datasets to demonstrate the effectiveness of our method.

\section{RELATED WORK}
\label{RELATED WORK}
\subsection{Surface Reconstruction}

The NeRF-based method \cite{NeRF} employs a multi-layer perceptron neural network to learn the mapping from 2D image pixel rays to the colors and densities of 3D spatial points, synthesizing images through volume rendering. It iteratively updates the network parameters by optimizing the discrepancies with the real image, achieving highly accurate 3D surface reconstruction. 
Representative neural implicit surface methods \cite{VolSDF, NeuS} focus on neural implicit surface representations as the core, combining the geometric priors of traditional SDF with the flexible representation of neural radiance fields. This addresses the issues of surface ambiguity and multiple geometric assumptions in NeRF-based methods, achieving sub-millimeter accuracy in the reconstruction of complex topologies. Another methods \cite{UNISURF, NeRF++} surpasses the architectural limitations of traditional NeRF by employing spatial decoupling and rendering paradigm fusion. This breakthrough enables the reconstruction of open scenes and weakly textured regions, overcoming the bottlenecks faced by traditional methods. However, these methods consume significant memory and take considerable time, impacting the realistic representation of the scene and lacking real-time performance.

To address the above problems, 3DGS \cite{3DGS} simulates the scene by constructing a set of learnable Gaussian functions. Each Gaussian element characterizes local geometry and appearance features through its parameters, including position, covariance, opacity, and color, and dynamically adjusts them to accommodate the complex and evolving scene structure. Representative explicit surface methods \cite{SuGaR, GOF} directly extract explicit geometric structures in 3D space and transform implicit Gaussian fields into explicit geometric representations. This provides a structured basis for downstream tasks, including scene editing and physics simulation. Meanwhile, 2DGS \cite{2DGS} unfolds the 3D Gaussian function into a 2D Gaussian disc, providing view-consistent geometric information to facilitate surface interior modeling. Additionally, 2DGS integrates depth distortion correction and normal consistency verification to enhance the accuracy of surface reconstruction. However, despite the advancements made by these 3DGS-based methods, they still grapple with issues of depth bias and inadequate geometric consistency across multiple viewpoints.

\subsection{Geometric Regularization}  

In 3DGS-based scene reconstruction, the geometric regularization strategy serves as the core technical method to overcome the bottleneck of reconstruction quality. Current studies focus on effectively addressing the limitations of traditional methods, particularly regarding implicit constraints and other aspects, through mechanisms like depth field optimization and surface normal constraints. 
One class of methods \cite{D2T, SIDGaussian} takes depth information as the core optimization objective and addresses the issues of depth jumping and multi-view mismatch in traditional methods through a global-local optimization strategy for depth prediction. This significantly enhances the geometric robustness of complex scenes. Another class of methods \cite{DyGASR, GeoTexDensifier} emphasizes the explicit optimization of geometric representations, starting from the underlying parameters of these representations. By combining explicit physical constraints, these methods tackle problems such as surface breakage and anomaly interference, enabling high-fidelity geometric reconstruction. However, the absence of inter-view geometric coherence regularization in existing frameworks fundamentally limits their multiview reconstruction fidelity. As a result, the normal vector of the Gaussian may perform well in a single-view direction but deviate from the true surface geometry in other views, failing to ensure consistency across all views.

\section{METHOD}
\label{METHOD}

\begin{figure*}
    \centering
    \includegraphics[width=1\linewidth]{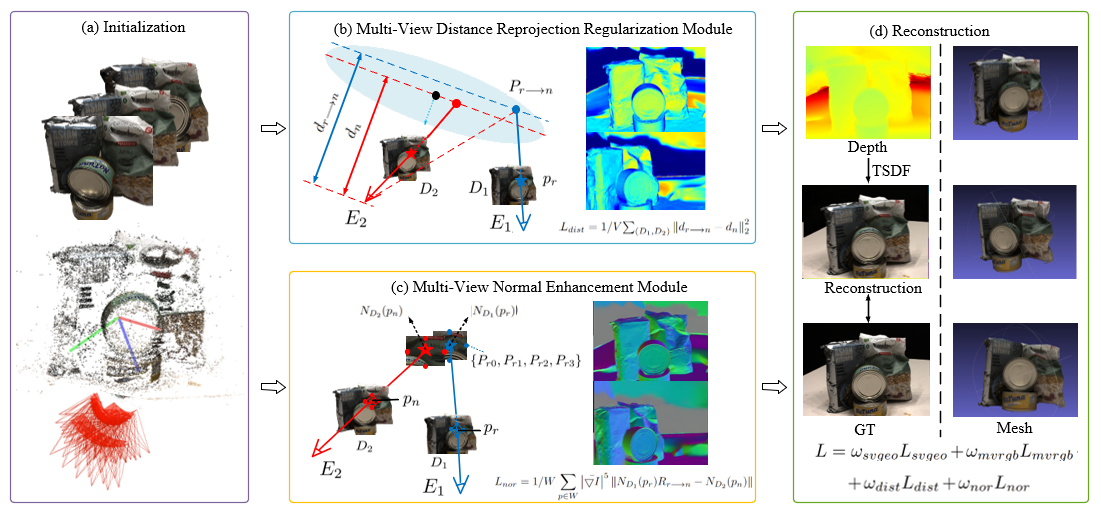}
    \caption{Pipeline. (a) COLMAP is used to initialize the input nearby views in order to obtain the point cloud. (b) The multi-view distance reprojection regularization module ensures consistent geometric distances by calculating the distances of nearby viewpoints to the same plane and generating losses. (c) The multi-view normal enhancement module projects the same plane corresponding to nearby viewpoints into 3D space, computes their normals, and generates losses to reduce the deviation in normal estimation.}
    \label{fig:pipeline}
\end{figure*}

As shown in Fig. \ref{fig:pipeline}, given a set of input views, we utilize COLMAP to generate the point cloud and camera poses, each view corresponds to a viewpoint $E_{i}$. When representing a Gaussian distribution of the same object plane across these two nearby views, we employ camera ray direction $R_{ri}$ and depth information ${D}_{ri}$ to back-project the pixel $p_r$ into the reference camera space, yielding the 3D coordinate $P_r$. This coordinate is then transformed using the transformation matrix $\pi$ to the camera coordinate system of $E_2$, resulting in $P_{r\longrightarrow n}$. By multiplying the normal $v$ with the points $P_{r\longrightarrow n}$, we obtain projected distances, which are used to construct the multi-view distance reprojection regularization loss $L_{dist}$ for optimizing geometric distances. For nearby viewpoints, we construct pixel correspondences $p_r\longleftrightarrow p_n$ using a pre-computed homography matrix. For each pixel $p_r$ in depth map $D_1$, we consider its local neighborhood, we sample four neighboring points using a fixed template to form a candidate set $\left \{ p_{r0}, p_{r1}, p_{r2}, p_{r3} \right \} $, and project these to obtain 3D points $\left \{ P_{r0}, P_{r1}, P_{r2}, P_{r3} \right \} $. We fit a plane to these points to derive normals $N_{D_1}(p_r)$ and $N_{D_2}(p_n)$. A multi-view normal enhancement loss $L_{nor}$ is then constructed to optimize the geometric normals. The mesh reconstruction pipeline initially generates depth maps for each viewpoint, which are subsequently integrated into a volumetric TSDF representation using a fusion algorithm.

\subsection{Preliminaries}

\textbf{3D Gaussain Splatting.} 3DGS \cite{3DGS} comprises anisotropic 3D Gaussian primitives, and each Gaussian distribution achieves pixel shading by utilizing ray casting. The Gaussian $G(x)$ is as follows:

\begin{equation}
  \label{eq:Gx}
  G(x)=exp({-\frac{1}{2}(x-\mu )^T\Sigma ^{-1} (x-\mu )} )
\end{equation}
where $\mu$ represents the point cloud position and $\Sigma$ denotes the covariance, the 3D Gaussian projection is rendered into 2D space. The following covariance matrix $\Sigma'$ in camera coordinates:

\begin{equation}
  \label{eq:Sigma'}
 \Sigma' =JW \Sigma W^TJ^T
\end{equation}
where $J$ is the Jacobian matrix of the affine approximation of the projection transform and $W$ is the observation transformation matrix. Given the scaling matrix $S$ and the rotation matrix $R$, the optimized covariance matrix $\Sigma$ can be derived:
\begin{equation}
  \label{eq:Sigma}
 \Sigma =RSS^TR^T
\end{equation}

In the rendering process, each 3D Gaussian is associated with opacity $\alpha_i$:
\begin{equation}
  \label{eq:alpha_i}
  \alpha_i=\sigma_ie^{-\frac{1}{2}({p}-\mu_i)^T\sum^{-1}({p}-\mu_i)}
\end{equation}
where $\sigma_i$ is the density of Gaussians and $p$ represents the pixel's coordinates, $\mu_i$ denotes the coordinates of the 3D Gaussian when projected onto the 2D image plane. Subsequently, we perform point-based volume rendering of the color $c({p})$:

\begin{equation}
  \label{eq:c(p)}
  c({p})=\sum_{i\in N}T_i\alpha_ic
\end{equation}
where $T_i$ is the transmittance defined by $T_i=\Pi_{j=1}^{i-1}(1-\alpha_{j})$ and $c$ represents the view-dependent color.
 
\textbf{PGSR.} The 3D Gaussians are converted into 2D flat Gaussians, which serve as planes for subsequent optimization, the shortest axis of the Gaussian sphere serves as the normal of the Gaussian plane. PGSR \cite{PGSR} employs a method akin to 3DGS to compute colors by first acquiring a normal map $N$ and a distance map $\mathcal{D}$, and then generating a depth map $D(p)$ based on these two maps. The normal map $N$ is rendered using $\alpha-$ blending as follows:

\begin{equation}
  \label{eq:N}
  N=\sum_{i\in N}R_{c}^{T}n_i\alpha_i\prod_{j=1}^{i-1} (1-\alpha_j)  
\end{equation}
where the rotation matrix from camera coordinates to global coordinates is expressed as $R_{c}$. $d_i=(R_c^T(\mu _i-T_c))R_c^Tn_i^T$ represents the distance from the plane to the camera center. $T_c$ denotes the camera center in the world coordinate system. The shortest axis of the ellipse is defined as the normal to each Gaussian $n_i$. The distance map $\mathcal{D}$ is rendered using $\alpha-$blending as follows:

\begin{equation}
  \label{eq:DD}
   \mathcal{D} =\sum_{i\in N} d_i\alpha _i\prod_{j=1}^{i-1} (1-\alpha_j) 
\end{equation}

The depth map $D(p)$ can be obtained from the normal and distance map:

\begin{equation}
  \label{eq:D}
   D(p)=\frac{\mathcal{D} }{N(p)K^{-1}\tilde{p} } 
\end{equation}

PGSR proposes single-view normal loss $L_{svgeo}$ to address the issue of local overfitting, and multi-view consistency constraint $L_{mvrgb}$ to ensure the global consistency of these structures.

\subsection{Multi-View Distance Reprojection Regularization Module}

PGSR ensures unbiased depth estimation by dividing the distance map by the normal map and adjusting the ray accumulation weights. However, this method heavily relies on the accuracy of normal and planar estimations. Additionally, due to the inherent limitations of COLMAP, the distance of camera viewpoints from the same surface may vary across different viewpoints, resulting in inaccurate distance estimation. To address this issue and mitigate distance inconsistency, we propose a multi-view distance reprojection regularization module. This module enforces a projection constraint between the distances of two views, ensuring geometric consistency and alignment. By leveraging multi-view geometric relationships, our method achieves uniform and precise distance estimation, thereby enhancing the robustness and reliability of the reconstruction process.

We select the reference viewpoint $E_1$ and the nearby viewpoint $E_2$ as input data. For any pixel $p_r$ in the reference viewpoint $E_1$, it can be mapped to the corresponding pixel $p_n$ in the nearby viewpoint $E_2$ through the calculation of the obtained homography matrix $H_r$:

\begin{equation}
  \label{eq:pn}
  p_{n}=H_r \times p_r 
\end{equation}

\begin{equation}
  \label{eq:Hrn}
  H_{rn}=K_{n}(R_{rn}-\frac{T_{rn}n_{r}^{T} }{d_{r}}  )K_{r}^{-1} 
\end{equation}

After finding the co-view region, an explicit association model between the camera center and the object surface geometry is established. In reference view $D_1$, we back-project a pixel $p_r$ into the camera space of the reference viewpoint $E_1$ by utilizing the direction of the camera ray passing through it and its depth information. During this process, each ray direction $R_{ri}$ vector is scaled by its corresponding depth value ${D}_{ri}$, resulting in the 3D coordinates that are obtained from the observation point after applying depth scaling along the ray direction. These coordinates represent the corresponding 3D position $P_r$ of the pixel $p_r$ within the reference camera coordinate system:

\begin{equation}
  \label{eq:Pri}
  P_{ri}=R_{ri}\cdot {D}_{ri}
\end{equation}
where $R_{ri}$ represents the direction vector of the $i$-th ray $(i= 1,2,3,4)$, ${D}_{ri}$ is the depth value from the observation point to the point in the direction of the $i$-th ray. Next, we utilize the camera's extrinsic to transform it into the world coordinate system, and subsequently apply the extrinsic of the viewpoint $E_2$ to convert it into the camera coordinate system of $E_2$, yielding $P_{r \longrightarrow n}$:

\begin{equation}
  \label{eq:Prn}
  P_{r\longrightarrow n}=\pi P_r
\end{equation}
where $\pi$ is the transformation of $P_r$ from the current camera coordinate system to the world coordinate system and then to the camera coordinate system of another viewpoint. Then, as shown in Fig. \ref{fig:pipeline} (b), based on the geometric relationship, the vector $P_{r \longrightarrow n}$ is dot-multiplied with the normal $\vec{v} $ corresponding to the point $p_n$ to obtain the projected distance $P_{r \longrightarrow n}$. Subsequently, this projected distance $d_{r \longrightarrow n}$ is compared with the original distance $d_n$ corresponding to the point $p_n$ to compute the multi-view distance reprojection regularization loss $ L_{dist}$:

\begin{equation}
  \label{eq:L}
 L_{dist}= 1/V{\textstyle \sum_{(D_1,D_2)}} \left \| d_{r \longrightarrow n}-d_n \right \|^2_2  
\end{equation}

\subsection{Multi-View Normal Enhancement Module}

Single-view constraints alone are insufficient for accurate normal estimation because Gaussians optimized under these constraints may deviate from the true surface geometry from different perspectives. To ensure consistent and accurate normals, we incorporate a multi-view normal enhancement module. This module aligns estimated normals with the object's surface across all views, reducing deviations and enhancing reconstruction accuracy. Our method improves robustness and ensures a more faithful geometry representation, especially for complex surfaces and textureless regions

Given two input views $E_1$ and $E_2$, pixel-level correspondences $p_r\longleftrightarrow p_n$ are first established based on the precomputed inter-view monoclinicity matrix. For the pixel $p_r$ in the depth map $D_1$ of the reference viewpoint $E_1$, the local neighborhood region centered on this pixel is extracted. Nearby points are then selected to construct the candidate point set $\left \{ p_{r0}, p_{r1}, p_{r2}, p_{r3} \right \} $. Subsequently, four points $\left \{ P_{r0}, P_{r1}, P_{r2}, P_{r3} \right \} $ can be obtained by projecting them into the 3D space of the reference viewpoint, based on the ray direction $R_{ri}$ and depth information ${D}_{ri}$, respectively, as shown in Equation \ref{eq:Pri}.

These 3D spatial points are utilized for plane fitting to derive the normal direction $N_{D_1}(p)$ and $N_{D_2}(p)$ of the local geometric surface:

\begin{equation}
  \label{eq:N_d(p)}
  N_{D}(p_{r})=\frac{(P_{r1}-P_{r0})\times (P_{r3}-P_{r2})}{\left |(P_{r1}-P_{r0}) \times (P_{r3}-P_{r2}) \right | }   
\end{equation}

Finally, we construct multi-view normal enhancement loss $L_{nor}$, and the parameters are optimized through a micro-renderable mechanism: 
\begin{equation}
  \label{eq:Lnor}
  L_{nor}= 1/W\sum_{p\in W}\left |\bar{\bigtriangledown I}   \right |^5\left \| N_{D_1}(p_r)R_{r\longrightarrow n}- N_{D_2}(p_n)\right \|  
\end{equation}

The module forces the local geometric features from different viewpoints to converge to a uniform geometric manifold, thereby significantly improving the geometric consistency in normal estimation across viewpoints.

\subsection{Loss Function}

In the final surface reconstruction process, we integrate PGSR with respect to both single-view normal loss $L_{svgeo}$ and multi-view consistency constraint $L_{mvrgb}$ to optimize Gaussian geometry. We configure different weights for each loss component accordingly:
\begin{equation}
  \label{eq:L}
  L=\omega_{svgeo}L_{svgeo}+\omega_{mvrgb}L_{mvrgb} +\omega_{dist} L_{dist}+\omega_{nor} L_{nor}
\end{equation}

\section{EXPERIMENTS}
\label{EXPERIMENTS}
\begin{figure*}
    \centering
    \includegraphics[width=1\linewidth]{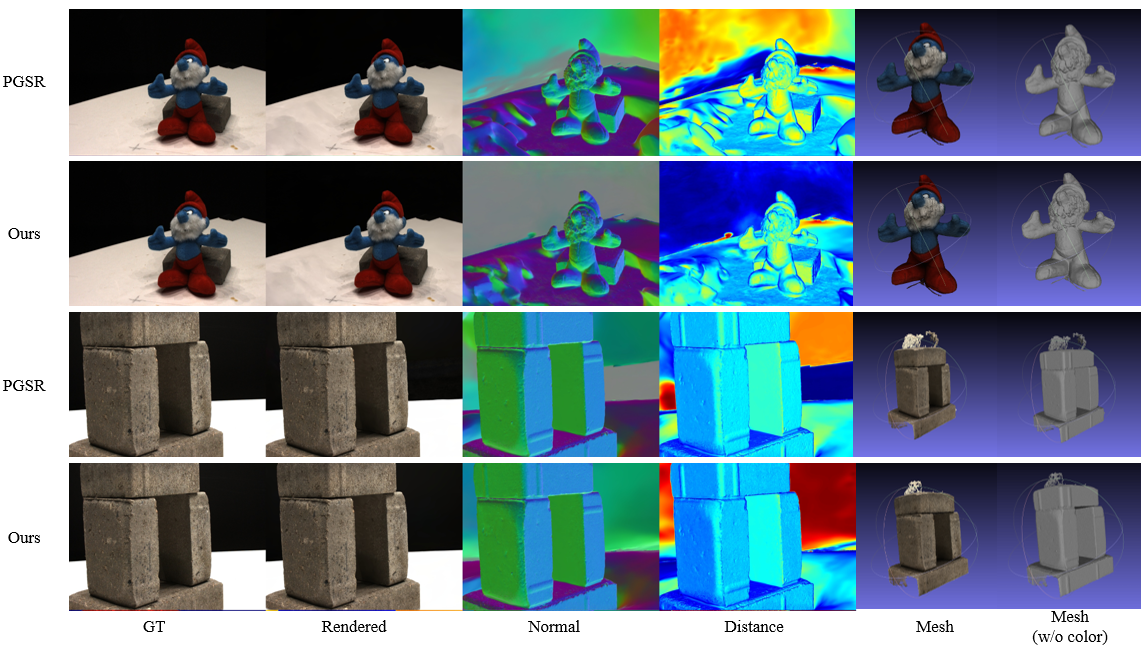}
    \caption{Qualitative comparison on the DTU dataset. The odd rows of the figure display the reconstruction results of PGSR, whereas the even rows exhibit the reconstruction results obtained by our method. The first column depicts the ground truth (GT), and the second column shows the corresponding reconstruction outcomes. The third column features the normal map, and the fourth column presents the distance map. The fifth and sixth columns are mesh maps with and without color, respectively}
    \label{fig:comp-DTU}
\end{figure*}

\begin{figure*}
    \centering
    \includegraphics[width=1\linewidth]{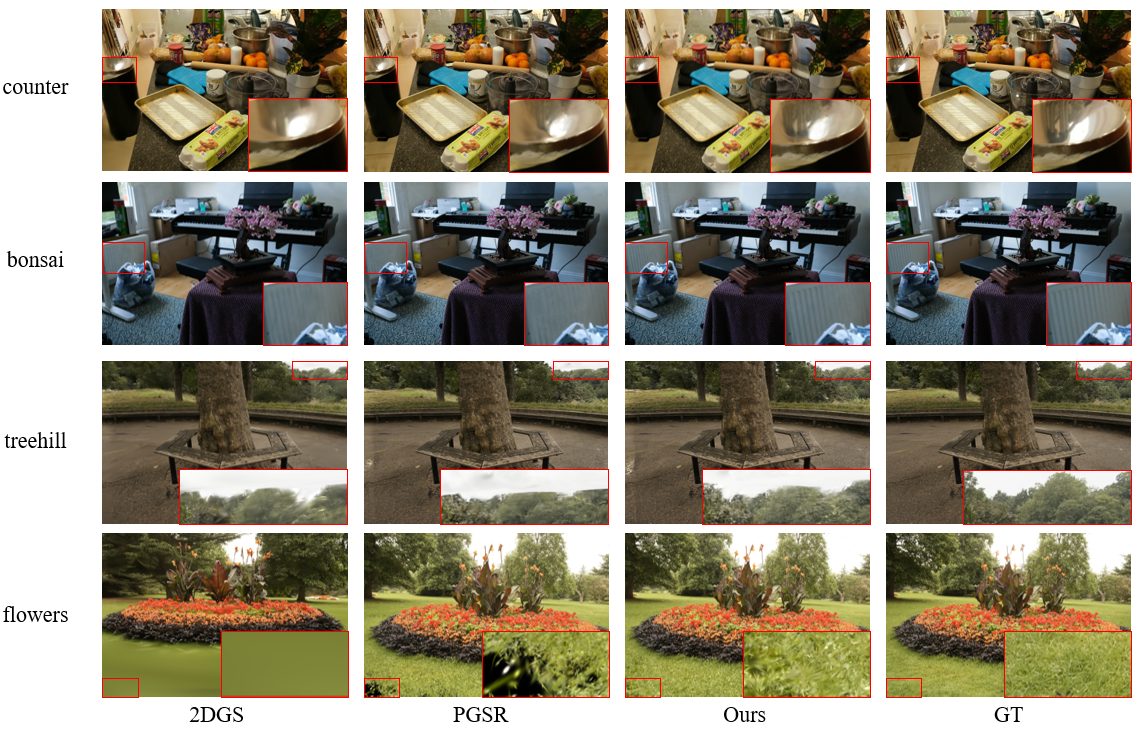}
    \caption{Qualitative comparison on the Mip-NeRF360 dataset. We compare our method with 2DGS and PGSR. Lines 1-2 display the rendering results for indoor scene, while lines 3-4 present the rendering results for outdoor scene.}
    \label{fig:comp-mip}
\end{figure*}

\subsection{Experiment Details}
\label{sec:Experiments Details}
\textbf{(1) Dataset.}
To maintain consistency with PGSR \cite{PGSR}, we have chosen 15 scenes from the DTU \cite{DTU} dataset for our surface reconstruction task. The DTU dataset is a comprehensive multi-view stereo vision dataset that includes 128 scenes, all captured in a controlled laboratory setting. Each scene is recorded from either 49 or 64 camera angles under seven different lighting conditions, yielding RGB images.

In addition, we synthesize novel views on the Mip-NeRF360 \cite{Mip-NeRF360} dataset. The Mip-NeRF360 dataset is specifically designed for borderless rendering and comprises a total of 9 scenes, including 5 outdoor scenes and 4 indoor scenes. 

\textbf{(2) Evaluation Metric.}
To assess the effectiveness of surface reconstruction on the DTU dataset, we utilize the Chamfer Distance (CD). This metric measures the similarity between two point clouds by computing the average distance from each point in one cloud to its nearest neighbor in the other. A lower CD value signifies higher similarity between the point clouds.

To evaluate novel view synthesis, we use PSNR, SSIM \cite{SSIM}, and LPIPS \cite{LPIPS}. PSNR quantifies reconstruction quality. SSIM evaluates similarity based on brightness, contrast, and structure. LPIPS assesses similarity using deep features, aligning with human vision. Cell colors indicate \colorbox{tabfirst}{top}, \colorbox{tabsecond}{second}, and \colorbox{tabthird}{third} best performances.

\textbf{(3) Implementation Details.}
We utilize the default parameters of PGSR in conjunction with the densification module introduced in AbsGS \cite{AbsGS}. Initially, we train solely the 3D Gaussian model for 7,000 iterations. Subsequently, in an additional 23,000 iterations, we replace the multi-view geometric consistency module in PGSR with our proposed multi-view distance reprojection regularizaion module. Furthermore, we refine the results by applying our Multi-View Normal Enhancement strategy, maintaining the same training strategy as PGSR. The weight assigned to the Multi-View Reprojection Distance Error module is 0.03, with an upper threshold of 0.03. Meanwhile, the weight of the Normal Enhancement module is set to 0.015, and its upper threshold is 0.52. Surface geometry is ultimately reconstructed through TSDF field iso-surface extraction. All experiments conducted in this paper were run on Nvidia RTX 4090 GPUs.

\begin{table*}[ht]
\centering
\caption{Quantitative comparison on the DTU Dataset. We report the errors of our method, in comparison to NeRF-based and 3DGS-based methods, on the CD metric. Among all the SOTA methods, our method exhibits the best performance and requires comparable time to PGSR(DS).}
\vspace{-0.2cm}
\resizebox{.98\textwidth}{!}{
\begin{tabular}{@{}llcccccccccccccccclcc}
\hline
 \multicolumn{3}{c}{} & 24 & 37 & 40 & 55 & 63 & 65 & 69 & 83 & 97 & 105 & 106 & 110 & 114 & 118 & 122 & & Mean & Time \\ \cline{4-18} \cline{20-21}
 & VolSDF~\cite{VolSDF} & &  1.14 &  1.26 &  0.81 & 0.49 & 1.25 &  0.70 &  0.72 &  1.29 &  1.18 &  \cellcolor{tabthird}0.70 & 0.66 & 1.08 &  0.42 &   0.61 &  0.55 & & 0.86 & $>\text{12h}$\\
 & NeuS~\cite{NeuS} & &  1.00 & 1.37 & 0.93 &  0.43 & 1.10 &   0.65 &    0.57 &  1.48 &   1.09 &  0.83 &  \cellcolor{tabthird}0.52 &  1.20 &  0.35 & \cellcolor{tabthird}0.49 &  0.54 & &  0.84 & $>\text{12h}$\\
 & Neuralangelo~\cite{Neuralangelo} & &  \cellcolor{tabthird}0.37 & \cellcolor{tabthird}0.72 &  \cellcolor{tabfirst}0.35 & \cellcolor{tabfirst}0.35 &  \cellcolor{tabthird}0.87 & \cellcolor{tabfirst}0.54 &  \cellcolor{tabthird}0.53 &  1.29 &  \cellcolor{tabthird}0.97 &  0.73 &  \cellcolor{tabsecond}0.47 & \cellcolor{tabthird}0.74 &  \cellcolor{tabthird}0.32 & \cellcolor{tabsecond}0.41 &  \cellcolor{tabthird}0.43 & & \cellcolor{tabthird}0.61 & $>\text{12h}$\\ 
 \cline{2-2} \cline{4-18} \cline{20-21}
 &  SuGaR~\cite{SuGaR} & & 1.47 & 1.33 & 1.13 & 0.61 & 2.25 & 1.71 & 1.15 & 1.63 & 1.62 & 1.07 & 0.79 & 2.45 & 0.98 & 0.88 & 0.79 & & 1.33  & \cellcolor{tabthird}1h \\
 & 2DGS~\cite{2DGS} &&  0.48 & 0.91 &  \cellcolor{tabthird}0.39 &  0.39 &  1.01 &  0.83 &  0.81 &  1.36 &  1.27 &  0.76  &  0.70 &  1.40 &   0.40 &   0.76 &  0.52 &&   0.80 & \cellcolor{tabfirst}0.32h \\
 & GOF~\cite{GOF} & &  0.50 & 0.82 &\cellcolor{tabsecond}0.37 & \cellcolor{tabthird}0.37 & 1.12 &  0.74 & 0.73 &  \cellcolor{tabthird}1.18 & 1.29 & \cellcolor{tabsecond}0.68 & 0.77 &  0.90 & 0.42 & 0.66 & 0.49 &&  0.74 & 2h\\
 & PGSR(DS)~\cite{PGSR} & &  \cellcolor{tabsecond}0.36 & \cellcolor{tabfirst}0.55 & \cellcolor{tabthird}0.39 & \cellcolor{tabsecond}0.36 & \cellcolor{tabsecond}0.78 & \cellcolor{tabsecond}0.58 & \cellcolor{tabfirst}0.49 & \cellcolor{tabsecond}1.07 & \cellcolor{tabsecond}0.63 &  \cellcolor{tabfirst}0.59 & \cellcolor{tabsecond}0.47 & \cellcolor{tabsecond}0.54 & \cellcolor{tabfirst}0.30 &  \cellcolor{tabfirst}0.37 &  \cellcolor{tabsecond}0.34 && \cellcolor{tabsecond}0.52  & \cellcolor{tabsecond}0.5h\\

   \cline{2-2} \cline{4-18} \cline{20-21}
  & Ours(DS) & & \cellcolor{tabfirst}0.34 & \cellcolor{tabsecond}0.56 & \cellcolor{tabsecond}0.37 &  \cellcolor{tabfirst}0.35 &  \cellcolor{tabfirst}0.77 &  \cellcolor{tabthird}0.62 &\cellcolor{tabsecond}0.50 &  \cellcolor{tabfirst}1.04 & \cellcolor{tabfirst}0.62 & \cellcolor{tabfirst}0.59 & \cellcolor{tabfirst}0.46 & \cellcolor{tabfirst}0.48 & \cellcolor{tabsecond}0.31 & \cellcolor{tabfirst}0.37 &  \cellcolor{tabfirst}0.33 &&  \cellcolor{tabfirst}0.51  &\cellcolor{tabsecond}0.5h\\
 \hline
\end{tabular}
}
\label{tab:dtu_result}
\vspace{-0.1cm}
\end{table*}

\begin{table*}[ht]
\centering
\caption{Quantitative results on Mip-NeRF360 dataset. Our method achieves the highest SSIM and the lowest LPIPS values on average across various scenes. In particular, our PSNR metric outperforms all others in outdoor scenes.}
\vspace{-0.2cm}
\begin{tabular}{l ccc ccc ccc}
 \hline
 & \multicolumn{3}{c}{Outdoor Scene} & \multicolumn{3}{c}{Indoor scene} & \multicolumn{3}{c}{Average scene}\\ 
 & PSNR~$\uparrow$ & SSIM~$\uparrow$ & LPIPS~$\downarrow$ & PSNR~$\uparrow$ & 
SSIM~$\uparrow$ & LIPPS~$\downarrow$ & PSNR~$\uparrow$ & 
SSIM~$\uparrow$ & LIPPS~$\downarrow$\\
\hline
NeRF ~\cite{NeRF} & 21.46 & 0.458 & 0.515 & 26.84 & 0.790 & 0.370 & 24.15 & 0.624 & 0.443 \\
Deep Blending ~\cite{Deep Blending} &  21.54 & 0.524 & 0.364 & 26.40 & 0.844 & 0.261 & 23.97 & 0.684 & 0.313 \\
Instant NGP ~\cite{iNGP} & 22.90 & 0.566 & 0.371 & 29.15 & 0.880 & 0.216 &  26.03 & 0.723 & 0.294 \\
Mip-NeRF360 ~\cite{Mip-NeRF360} & 24.47 & 0.691 & 0.283 & \cellcolor{tabfirst}31.72 & 0.917 & 0.180 &  \cellcolor{tabfirst}28.10 &  0.804 & 0.232 \\
NeuS ~\cite{NeuS} & 21.93 & 0.629 & 0.600 &  25.10 & 0.789 & 0.319 & 23.74 & 0.720 & 0.439 \\
\hline
3DGS ~\cite{3DGS} &  24.24 & 0.705 &  0.283 & \cellcolor{tabsecond}30.99 & 0.926 & 0.199 &  27.24 & 0.803 & 0.246 \\
SuGaR ~\cite{SuGaR} &  22.76 &  0.631 &  0.349 &  29.44 &  0.911 &  0.216 & 26.10 &   0.771 &   0.283  \\
2DGS ~\cite{2DGS} &  24.33 & 0.709 & 0.284 & 30.39 & 0.923 & 0.183 & 27.03 & 0.804 & 0.239 \\
GOF ~\cite{GOF} & \cellcolor{tabthird}24.76 & \cellcolor{tabthird}0.742 & \cellcolor{tabthird}0.225  & \cellcolor{tabthird}30.80 & \cellcolor{tabthird}0.928 & \cellcolor{tabthird}0.167  & \cellcolor{tabsecond}27.78 & \cellcolor{tabthird}0.835 &  \cellcolor{tabthird}0.196 \\
PGSR ~\cite{PGSR} &  \cellcolor{tabsecond}24.82 &  \cellcolor{tabsecond}0.753 &  \cellcolor{tabsecond}0.203 & 30.28 &   \cellcolor{tabsecond}0.930 &   \cellcolor{tabsecond}0.158 & 27.55 &   \cellcolor{tabsecond}0.842 &   \cellcolor{tabsecond}0.181  \\
\hline
Ours & \cellcolor{tabfirst}24.92 & \cellcolor{tabfirst}0.755 & \cellcolor{tabfirst}0.198 & 30.45 &  \cellcolor{tabfirst}0.932 &   \cellcolor{tabfirst}0.154  & \cellcolor{tabthird}27.69 &   \cellcolor{tabfirst}0.844 &   \cellcolor{tabfirst}0.176 \\
 \hline
\end{tabular}
\label{tab:mipnerf360}
\end{table*}

\subsection{Surface Reconstruction}

We perform surface reconstruction on the DTU dataset and compare the results with current state-of-the-art baseline methods to evaluate the effectiveness of our approach. It is worth noting that the PGSR paper mentions two types of metrics: those derived directly from original sampling, and those obtained by training on images downsampled (DS) to half their original size. The PGSR code, publicly available on GitHub, involves downsampling images and adjusting relevant parameters. After adjusting the parameters, we reproduced the code and accordingly derived the PGSR metrics presented in Fig. \ref{fig:comp-DTU}. As for the quality metrics of other methods, they are sourced from PGSR \cite{PGSR}. To align with PGSR, our method similarly includes downsampling the images.

As shown in Table I, we compare our method with NeRF-based and 3DGS-based methods. NeRF-based methods (VolSDF, NeuS, Neuralangelo) have poorer CD values and longer rendering times ($>$12 hours). 3DGS-based methods (SuGaR, 2DGS, GOF, RaDe-GS, PGSR(DS)) are faster and perform better in CD metrics. Our method achieves the lowest average CD value, with a 0.01 reduction compared to the baseline, while maintaining similar rendering time.

As shown in Fig. \ref{fig:comp-DTU}, our method produces more stable normal maps than PGSR, with greater uniformity in the background. Our distance maps also capture surface unevenness effectively through color variations, whereas PGSR lacks precision and renders the sculpture in a uniform color.

\subsection{Novel View Synthesis}

We conduct novel view synthesis experiments on the Mip-NeRF360 dataset, and compare the results obtained with the current leading baseline techniques to verify the effectiveness of our method. The metrics utilized for quality assessment are sourced from the publicly available relevant papers.

Table \ref{tab:mipnerf360} presents the quantitative comparison results of our method against 10 other methods, including NeRF, Deepl Blending, Instant NGP, Mip-NeRF360, NeuS, 3DGS, SuGaR, 2DGS, GOF and PGSR, on the Mip-NeRF360 dataset. These comparisons are evaluated using metrics such as PSNR, SSIM, and LPIPS to demonstrate the effectiveness of our method. NeRF-based methods, due to their implicit characterization constraint limitations, often underperform in the field of novel view synthesis. To overcome this shortcoming, the 3DGS-based method has been improved, resulting in superior performance on this dataset in terms of PSNR and SSIM metrics compared to NeRF-based methods. Additionally, the LPIPS metrics for the 3DGS-based method are lower, indicating better results. Our method builds upon the 3DGS method with the inclusion of multi-view distance reprojection regularization module and multi-view normal enhancement module. When compared to the baseline method, our method achieves higher SSIM values of 0.843, respectively, while also reducing the LPIPS metrics by 0.005 in the average scene. In the outdoor scene, our method achieve the highest PSNR value of 24.92. 

Fig. \ref{fig:comp-mip} presents the results of novel view synthesis using our method alongside 2DGS and PGSR on the mip-nerf360 dataset, featuring six selected indoor and outdoor scenes. In the counter scene, our method successfully renders the shelves reflected in the mirrored lid, closely matching the GT. Conversely, in the bonsai scene, our method renders the heater pipes with greater clarity compared to the blurred results produced by the other two methods. PGSR exhibits artifacts in the treehill scenes, and it fails to fully populate the flowers scene, leaving black voids. Meanwhile, 2DGS struggles to accurately render the shape of the grass in the flowers scene, resulting in severe artifacts and blurriness.

\begin{figure}[h]
    \centering
\includegraphics[width=1\linewidth]{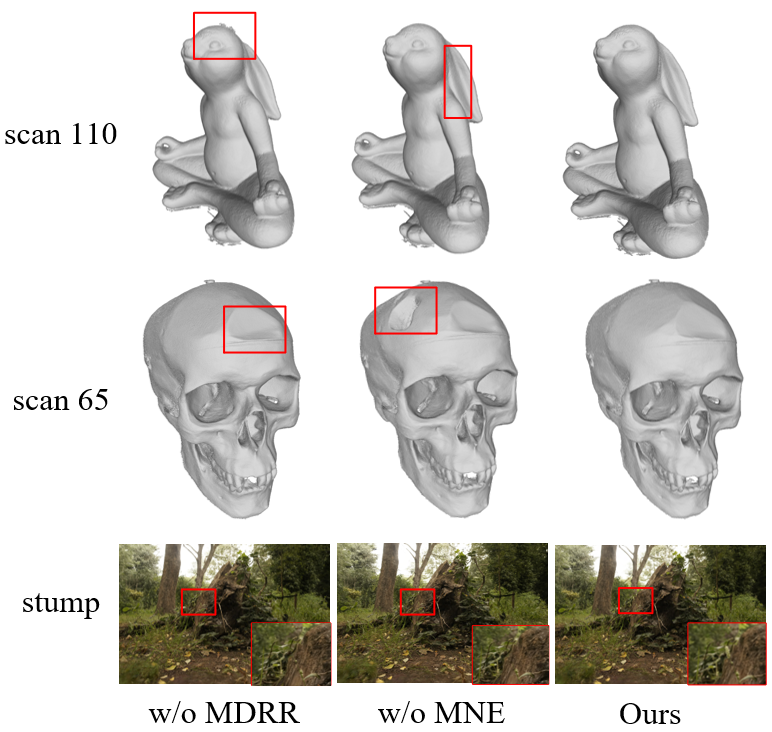}
\caption{Visualize ablation study results in all modules. We demonstrate the results of surface reconstruction and novel view synthesis when the MDRR module is removed, as well as when the MNE module is removed.}
\label{fig:abl}
\end{figure}

\subsection{Ablation Study}

\begin{table}[htbp]
\centering
\caption{Ablation study of CD on the DTU dataset and PSNR on the DTU and Mip-NeRF360 dataset.}
\label{tab:Ablation}
\begin{tabular}{lcc}
\hline
\multicolumn{1}{p{2cm}}{Model setting} &
\multicolumn{1}{p{2cm}}{scan110(CD)~$\downarrow$} &
\multicolumn{1}{p{2cm}}{stump(PSNR)~$\uparrow$}\\
\hline
w/o MDRR&0.54&27.24  \\
w/o MNE & 0.49&27.20  \\
\hline
Ours & 0.48 &27.24\\
\hline
\end{tabular}
\end{table}

As shown in Table \ref{tab:Ablation}, we conducted ablation experiments on the DTU and Mip-NeRF360 to assess the effectiveness of the multiview distance reprojection regularization (MDRR) module compared to the multiview normal enhancement module (MNE) in performing surface reconstruction. Specifically, the former model exhibits better performance in surface reconstruction, whereas the latter model achieves notable results in rendering.From the visualization results, the two modules we proposed have achieved good results in both detailed modeling of surface reconstruction and novel view synthesis, as shown in Fig. \ref{fig:abl}, the reconstruction and rendering results gradually become smoother and clearer from left to right.

To improve Gaussian sphere fitting in multi-view observation, we propose a multi-view distance projection regularization module. It uses light direction and depth from multiple viewpoints to ensure consistent distance information, reducing bias and improving reconstruction accuracy. As shown in Table \ref{tab:Ablation}, this module significantly improves CD by 0.06, indicating closer reconstruction to the real object surface.

The multi-view normal enhancement module addresses normal estimation inconsistencies in single-view systems. It integrates information from multiple viewpoints to ensure consistent local plane normals across all views, reducing bias and improving reconstruction accuracy. As shown in Table \ref{tab:Ablation}, this module improves PSNR by 0.04 and slightly increases CD, resulting in more faithful reconstruction and rendering.

\subsection{Limitations}
Though our method reconstructs smooth surfaces close to real objects, it's affected by initial point cloud and training set quality. Also, it's for small indoor/outdoor scenes, and projection errors spread in large-scale scenes.

\section{CONCLUSIONS}
For the reconstruction of small indoor and outdoor scenes, this paper addresses geometric inconsistency and surface fitting inaccuracy in 3DGS with multi-view conditions. Our proposed framework uses two innovative modules: multi-view distance reprojection regularization for consistent depth estimation and multi-view normal enhancement for normal surface alignment. Experiments on DTU and Mip-NeRF360 datasets confirm our method's efficacy in surface reconstruction and novel view synthesis.

\end{document}